\documentclass[onecolumn,journal]{IEEEtran}

\usepackage{arxiv}

\usepackage[utf8]{inputenc} 
\usepackage[T1]{fontenc}    
\usepackage{hyperref}       
\usepackage{url}            
\usepackage{booktabs}       
\usepackage{amsfonts}       
\usepackage{nicefrac}       
\usepackage{microtype}      

\usepackage{times}  
\usepackage{helvet} 
\usepackage{courier}  
\usepackage{graphicx} 

\usepackage{array,multirow}

\usepackage{algorithm}
\usepackage{algorithmic}

\newcounter{ALC@tempcntr}
\newcommand{\NEWCOMMENT}[1]{%
    \setcounter{ALC@tempcntr}{\arabic{ALC@rem}}
    \setcounter{ALC@rem}{1}
    \item {\footnotesize \it  //  #1}  
    \setcounter{ALC@rem}{\arabic{ALC@tempcntr}}
}%

\usepackage{enumitem}

\usepackage{wrapfig}

\usepackage{amsmath,amsthm,amsfonts}
\usepackage{bm}
\usepackage{amssymb}
\usepackage{xcolor}

\usepackage{subfigure}

\DeclareMathOperator{\bX}{\mathbf{X}}

\DeclareMathOperator{\bw}{\mathbf{w}}

\DeclareMathOperator{\fF}{\mathbb{F}}

\DeclareMathOperator{\gC}{\nabla C}

\DeclareMathOperator{\bnt}{\mathbf{n}^{(t)}}

\makeatletter
\newcommand*{\rom}[1]{\expandafter\@slowromancap\romannumeral #1@}
\makeatother

\newtheorem{theorem}{Theorem}
\newtheorem*{proof*}{proof}

\newtheorem{remark}{Remark}

\newcommand{\acr}{COPML} 

\title{A Scalable Approach for Privacy-Preserving Collaborative Machine Learning}

\author{
 Jinhyun So \\
  ECE Department \\
  University of Southern California (USC)\\
  \texttt{jinhyuns@usc.edu} \\
   \And
 Basak Guler \\
  ECE Department \\
  University of California, Riverside\\
  \texttt{bguler@ece.ucr.edu} \\
 \And
 A. Salman Avestimehr \\
  ECE Department \\
  University of Southern California (USC)\\
  \texttt{avestimehr@ee.usc.edu} \\
}

\begin{document}

\maketitle

\begin{abstract}
We consider a collaborative learning scenario in which multiple data-owners wish to jointly train a logistic regression model, while keeping their individual datasets private from the other parties. We propose  {\acr}, a fully-decentralized training framework that achieves scalability and privacy-protection simultaneously. The key idea of  {\acr} is to securely encode the individual datasets to distribute the computation load effectively across many parties and to  perform the training computations as well as the model updates in a distributed manner on the securely encoded data. 
We provide the privacy analysis of {\acr} and prove its convergence. Furthermore, we experimentally demonstrate that {\acr} can achieve significant speedup in training over the benchmark protocols. 
Our protocol provides strong statistical privacy guarantees against  colluding parties (adversaries) with unbounded computational power, while achieving up to $16\times$ speedup in the training time against the benchmark protocols. 
\end{abstract}

\section{Introduction}\label{sec:intro}
\vspace{-0.1cm}Machine learning applications can achieve significant performance gains by training on large volumes of data.
In many applications, the training data  is  distributed across multiple data-owners, such as patient records at multiple medical institutions, and furthermore contains sensitive information, e.g., genetic information, financial transactions, and geolocation information. Such settings give rise to the following key problem that is the focus of this paper: \emph{How can multiple data-owners jointly train a machine learning model while keeping their individual datasets private from the other parties?}

More specifically, we consider a distributed learning scenario in which $N$ data-owners (clients) wish to train a logistic regression model jointly without revealing information about their individual datasets to the other parties, even if up to $T$ out of $N$ clients collude.  
Our focus is on the semi-honest adversary setup, where the corrupted parties follow the protocol but may leak information in an attempt to learn the training dataset. 
To address this challenge, we propose a novel framework, {\acr}\footnote{COPML stands for collaborative privacy-preserving machine learning.}, that enables fast and privacy-preserving training by leveraging information and coding theory principles. 
{\acr} has three salient features: 
\vspace{-0.15cm}\begin{itemize}[leftmargin=0.4cm]

\item speeds up the training time significantly, by  distributing the  computation load effectively across a large number of parties,

\vspace{-0.1cm}\item advances the state-of-the-art privacy-preserving training setups by scaling to a large number of parties, as it can distribute the computation load effectively as more parties are added in the system,

\vspace{-0.1cm}\item utilizes coding theory principles to secret share the dataset and model parameters which can significantly reduce the communication overhead and the complexity of distributed training.

\vspace{-0.15cm}\end{itemize}

At a high level, {\acr} can be described as follows.
Initially, the clients secret share their individual datasets with the other parties, after which they carry out a secure multi-party computing (MPC) protocol to \emph{encode} the dataset. 
This encoding operation transforms the dataset into a \emph{coded} form that enables faster training and simultaneously guarantees privacy (in an information-theoretic sense). 
Training is performed over the encoded data via gradient descent. 
The parties perform the computations over the encoded data {\it as if they were computing over the uncoded dataset}. That is, the structure of the computations are the same for computing over the uncoded dataset versus computing over the encoded dataset. 
At the end of training, each client should only learn the final model, and no information should be leaked (in an information-theoretic sense) about the individual datasets or the intermediate model parameters, beyond the final model. 

We characterize the theoretical performance guarantees of {\acr}, in terms of  convergence, scalability, and privacy protection. Our analysis identifies a trade-off between privacy and parallelization, such that, each additional client can be utilized either for more privacy, by  protecting against a larger number of collusions $T$, or more parallelization, by  reducing the computation load at each client. 
Furthermore, we empirically demonstrate the performance of {\acr} by comparing it with  cryptographic benchmarks based on  secure multi-party computing (MPC)~\cite{yao1982protocols,ben1988completeness,beerliova2008perfectly,damgaard2007scalable}, 
that can also be applied to enable privacy-preserving machine learning tasks (e.g. see~\cite{nikolaenko2013privacy,gascon2017privacy,mohassel2017secureml,lindell2000privacy,dahl2018private,chen2019secure,SecureNN3Party,mohassel2018aby}). 
Given our focus on information-theoretic privacy, the most relevant MPC-based schemes for empirical comparison are the protocols from \cite{ben1988completeness} and  \cite{beerliova2008perfectly, damgaard2007scalable} based on Shamir's secret sharing~\cite{shamir1979share}. 
While several more recent works have considered MPC-based learning setups with information-theoretic privacy~\cite{SecureNN3Party,mohassel2018aby}, their constructions are limited to three or four parties. 

We run extensive experiments over the Amazon EC2 cloud platform to empirically demonstrate the  performance of {\acr}. We train a logistic regression model for image classification over the  CIFAR-10~\cite{krizhevsky2009learning} and GISETTE~\cite{NIPS2004_2728} datasets. The training computations are distributed to up to $N=50$ parties. We demonstrate that {\acr} can provide significant speedup in the training time against the state-of-the-art MPC baseline (up to $16.4\times$), while providing comparable accuracy to conventional logistic regression. 
This is primarily due to the parallelization gain provided by our system, which can distribute the workload effectively across many parties.

\noindent \textbf{Other related works.} 
Other than MPC-based setups, one can consider two notable approaches. The first one is Homomorphic Encryption (HE)~\cite{gentry2009fully}, which enables computations on encrypted data, and has been applied to  privacy-preserving machine learning ~\cite{gilad2016cryptonets,hesamifard2017cryptodl,graepel2012ml,yuan2014privacy,li2017privacy,Kim2018,8325493,logiscticHE}. 
The privacy protection of HE depends on the size of the encrypted data, and computing in the encrypted domain is computationally intensive. 
The second approach is differential privacy (DP), which is a noisy release mechanism to protect the privacy of personally identifiable information.
The main application of DP in machine learning is when the model is to be released publicly after training, so that individual data points cannot be backtracked from the released model \cite{chaudhuri2009privacy,shokri2015privacy,abadi2016deep,pathak2010multiparty,brendan2018learning,pmlr-v22-rajkumar12,jayaraman2018distributed}. 
On the other hand, our focus is on ensuring privacy during training, while preserving the accuracy of the model.
\vspace{-0.1cm}

\section{Problem Setting}
\vspace{-0.15cm}
We consider a collaborative learning scenario in which the training dataset is distributed across $N$ clients. Client  $j\in[N]$ holds an individual dataset denoted by a matrix $\mathbf{X}_j\in\mathbb{R}^{m_j\times d}$ consisting of $m_j$ data points with $d$ features, and the corresponding labels are given by a vector $\mathbf{y}_j\in\{0,1\}^{m_j}$. The overall dataset is denoted by  $\mathbf{X} =[\mathbf{X}_1^\top, \ldots, \mathbf{X}_N^\top]^\top$ consisting of $m\triangleq \sum_{j\in[N]}m_j$ data points with $d$ features, and corresponding labels $\mathbf{y} =[\mathbf{y}^\top_1, \ldots, \mathbf{y}^\top_N]^\top$, which consists of $N$ individual datasets each one belonging to a different client. The clients wish to jointly train a logistic regression model $\mathbf{w}$ over the training set $\mathbf{X}$ with labels $\mathbf{y}$, by minimizing a cross entropy loss function, 
\vspace{-0.1cm}\begin{equation}\label{cost}
C(\mathbf{w}) = \frac{1}{m}\sum_{i=1}^m \left ( -y_i \log \hat{y}_i  - (1-y_i) \log (1-\hat{y}_i)\right )    
\end{equation} 

\vspace{-0.3cm}\noindent
where $\hat{y}_i = g(\mathbf{x}_i \cdot \mathbf{w})\in (0,1)$ is the probability of label $i$ being equal to $1$, $\mathbf{x}_i$ is the $i^{th}$ row of matrix  $\mathbf{X}$, 
and $g(\cdot)$ denotes the sigmoid function $g(z) =1/(1+e^{-z})$. 
The training is performed through gradient descent, by updating the model parameters in the opposite direction of the gradient,
\vspace{-0.05cm}\begin{equation}\label{grad}
\mathbf{w}^{(t+1)}  =  \mathbf{w}^{(t)}-\frac{\eta}{m}\mathbf{X}^{\top}(g(\mathbf{X} \times \mathbf{w}^{(t)}) -\mathbf{y}) 
\end{equation}
where $\nabla C (\bw) = \frac{1}{m}\mathbf{X}^{\top}(g(\mathbf{X} \times \mathbf{w}) -\mathbf{y})$ is the gradient for \eqref{cost},  $\mathbf{w}^{(t)}$ holds the estimated parameters from iteration $t$,  $\eta$ is the learning rate, and function $g(\cdot)$ acts element-wise over the vector $\mathbf{X}\times \mathbf{w}^{(t)}$. 

During training, the clients wish to protect the privacy of their individual datasets from other clients, even if up to $T$ of them collude, where $T$ is  the {\it privacy parameter} of the system. There is no trusted party who can collect the datasets in the clear and perform the training. 
Hence, the training protocol should preserve the privacy of the individual datasets against any collusions between up to $T$ adversarial clients. 
More specifically, this condition states that the adversarial clients should not learn any information about the datasets of the benign clients beyond what can already be inferred from the adversaries' own datasets. 

To do so, client $j\in [N]$ initially secret shares its individual dataset $\mathbf{X}_j$  and  $\mathbf{y}_j$ with the other parties. 
Next, clients carry out a secure MPC protocol to encode the dataset by using the received secret shares. 
In this phase, the dataset $\mathbf{X}$ is first partitioned into $K$ submatrices  $\mathbf{X} = [\mathbf{X}_1^\top, \cdots, \mathbf{X}_K^\top]^\top$ for some $K \in \mathbb{N}$. 
Parameter $K$ characterizes the computation load at each client. Specifically, our system ensures that the computation load (in terms of gradient computations) at each client is equal to processing only $(1/K)^{th}$ of the entire dataset $\mathbf{X}$. 
The clients then encode the dataset by combining the $K$ submatrices together with some randomness to preserve privacy.
At the end of this phase, client $i\in [N]$ learns an encoded dataset $\widetilde{\mathbf{X}}_i$, whose size is equal to $(1/K)^{th}$ of the dataset $\mathbf{X}$. 
This process is only performed once for the dataset $\mathbf{X}$.

\begin{wrapfigure}{r}{0.5\textwidth}
\centering
\includegraphics[width=0.85\linewidth]{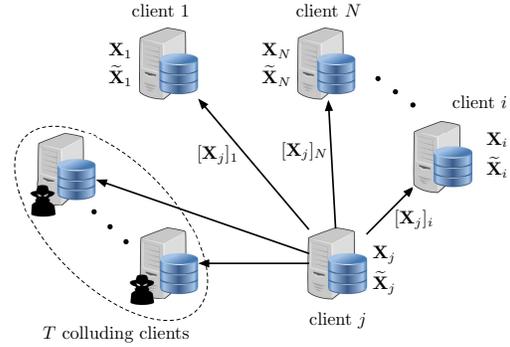}
\vspace{-0.2cm}\caption{The multi-client distributed training setup with $N$ clients. Client $j\in [N]$ holds a dataset  $\mathbf{X}_j$ with labels $\mathbf{y}_j$. At the beginning of training, client $j$ secret shares $\mathbf{X}_j$ and $\mathbf{y}_j$ to guarantee their information-theoretic privacy against any collusions between up to $T$ clients. The secret share of $\mathbf{X}_j$ and $\mathbf{y}_j$ assigned from client $j$ to client $i$ is represented by $[\mathbf{X}_j]_i$ and $[\mathbf{y}_j]_i$, respectively. 
}
\label{fig:masterless}
\vspace{-0.4cm}
\end{wrapfigure}
At each iteration of training, clients also encode the current estimation of the model parameters $\mathbf{w}^{(t)}$ using a secure MPC protocol, after which client $i\in [N]$ obtains the encoded model $\widetilde{\mathbf{w}}^{(t)}_i$. Client $i\in [N]$ then computes a local gradient   $\widetilde{\mathbf{X}}_i^\top g(\widetilde{\mathbf{X}}_i\times \widetilde{\mathbf{w}}^{(t)}_i)$ over the encoded dataset $\widetilde{\mathbf{X}}_i$ and encoded model  $\widetilde{\mathbf{w}}^{(t)}_i$. After this step, clients carry out another secure MPC protocol to decode the gradient  $\mathbf{X}^\top g(\mathbf{X} \times \mathbf{w}^{(t)})$ and update the model according to \eqref{grad}. As the decoding and model updates are performed using a secure MPC protocol, clients do not learn any information about the actual gradients or the updated model. In particular, client $i\in [N]$ only learns a secret share of the updated model, denoted by $[\mathbf{w}^{(t+1)}]_i$. 
Using the secret shares $[\mathbf{w}^{(t+1)}]_i$, clients $i\in [N]$ encode the model  $\mathbf{w}^{(t+1)}$ for the next iteration, after which client $i$ learns an encoded model $\widetilde{\mathbf{w}}^{(t+1)}_i$. 
Figure~\ref{fig:masterless} demonstrates our  system architecture.


\section{The {\acr} Framework}

\label{app:multiclient_framework}

\begin{wrapfigure}{R}{0.5\textwidth}
\vspace{-0.2cm}
\centering
\includegraphics[width=0.95\linewidth]{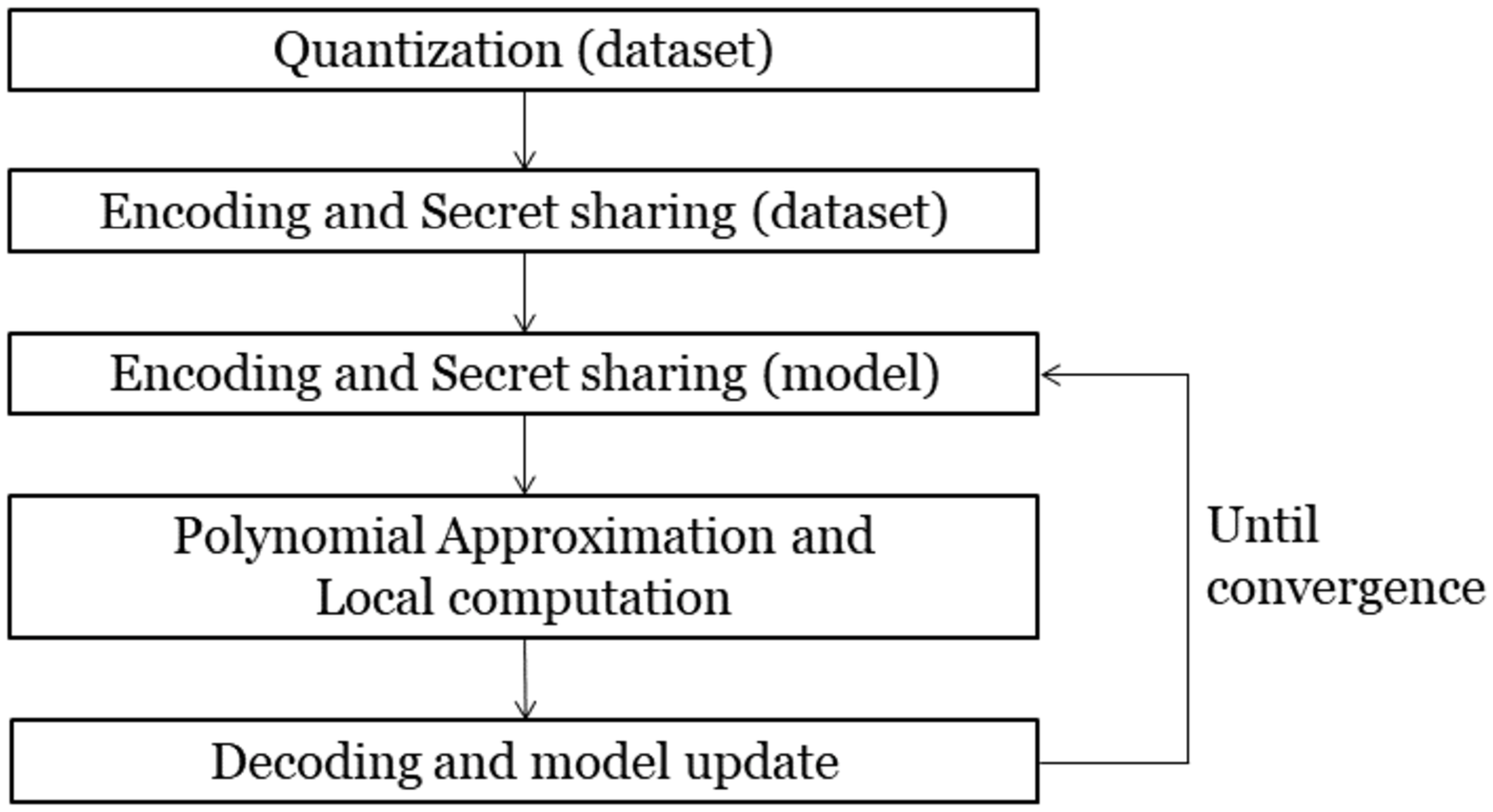}
\vspace{-0.2cm}
\caption{Flowchart of {\acr}. 
}
\vspace{-0.3cm}
\label{fig:blockdiagram}
\end{wrapfigure}

{\acr} consists of four main phases: quantization; encoding and secret sharing; polynomial approximation; decoding and model update, as demonstrated in Figure~\ref{fig:blockdiagram}. 
In the first phase, quantization, each client converts its own dataset from the real domain to finite field. In the second phase, clients create a secret share of their quantized datasets and carry out a secure MPC protocol to encode the datasets. At each iteration, clients also encode and create a secret share of the model parameters. In the third phase, clients perform local gradient computations over the encoded datasets and encoded model parameters by approximating the sigmoid function with a polynomial. Then, in the last phase, clients decode the local computations and update the model parameters using a secure MPC protocol. This process is repeated until the  convergence of the model parameters.  

\noindent{\bf Phase 1: Quantization.}
Computations involving secure MPC protocols are bound to finite field operations, which requires the representation of real-valued data points in a finite field $\fF$. 
To do so, each client initially quantizes its dataset from the real domain to the domain of integers, and then embeds it in a field $\mathbb{F}_p$ of integers modulo a prime $p$.  
Parameter $p$ is selected to be sufficiently large to avoid wrap-around in computations.
For example, in a $64$-bit implementation with the CIFAR-10 dataset, we select $p=2^{26}-5$.
The details of the quantization phase are provided in Appendix~\ref{app:quantization}.

\vspace{0.1cm}
\noindent{\bf Phase 2: Encoding and secret sharing.}
In this phase, client $j\in [N]$ creates a secret share of its quantized dataset $\mathbf{X}_j$ designated for each client $i\in [N]$ (including client $j$ itself). The secret shares are constructed via Shamir's secret sharing  with threshold $T$ \cite{shamir1979share}, to protect the privacy of the individual datasets against any collusions between up to $T$ clients. 
To do so, client $j$ creates a random polynomial, $h_j(z) = \bX_j + z \mathbf{R}_{j1}+ \ldots + z^T \mathbf{R}_{jT}$
where $\mathbf{R}_{ji}$ for $i\in [T]$ are i.i.d. uniformly distributed random matrices, and selects $N$ distinct evaluation points $\lambda_1, \ldots, \lambda_N$ from $\mathbb{F}_p$.
Then, client $j$ sends client $i\in[N]$ a secret share $[\mathbf{X}_j]_i\triangleq h_j(\lambda_i)$ of its dataset $\mathbf{X}_j$.
Client $j$ also sends a secret share of its labels $\mathbf{y}_j$ to client $i\in [N]$, denoted by $[\mathbf{y}_j]_i$. 
Finally, the model is initialized randomly within a secure MPC protocol between the clients, and at the end client $i\in[N]$ obtains a secret share $[\mathbf{w}^{(0)}]_i$ of the initial model $\mathbf{w}^{(0)}$. 


After obtaining the secret shares $[\mathbf{X}_j]_i$ for $j\in [N]$, clients $i\in[N]$ encode the dataset using a secure MPC protocol and transform it into a \emph{coded} form, which speeds up the training  by distributing the  computation load of gradient evaluations across the clients. 
Our encoding strategy utilizes Lagrange coding from \cite{yu2018lagrange}\footnote{Encoding of Lagrange coded computing is the same as a packed secret sharing \cite{franklin1992communication}.
}, which has been applied to other problems such as privacy-preserving offloading of a training task \cite{so2019codedprivateml} and secure federated learning \cite{so2020turbo}. 
However, we encode (and later decode) the secret shares of the datasets and not their true values. Therefore, clients do not learn any information about the true value of the dataset $\bX$ during the encoding-decoding process.

The individual steps of the encoding process are as follows. 
Initially, the dataset $\bX$ is partitioned into $K$ submatrices $\bX = [\mathbf{X}_1^\top, \ldots, \mathbf{X}_K^\top]^\top$ where $\bX_k \in \mathbb{F}^{\frac{m}{K}\times d}_{p}$ for $k\in[K]$. 
%
To do so, client $i\in[N]$ locally concatenates $[\mathbf{X}_j]_i$ for $j\in[N]$ and partitions it into $K$ parts,  $[\mathbf{X}_k]_i$ for $k\in[K]$. 
Since this operation is done over the secret shares, clients do not learn any information about the original dataset $\mathbf{X}$.
%
Parameter $K$ quantifies the computation load at each client, as will be discussed in  Section~\ref{sec:convergence}. 

The clients agree on $K+T$ distinct elements $\{\beta_k\}_{k\in[K+T]}$ and $N$ distinct elements~$\{\alpha_i\}_{i\in[N]}$  from $\mathbb{F}_p$ such that~$\{\alpha_i\}_{i\in[N]}\cap\{\beta_k\}_{k\in[K+T]}=\varnothing$. 
Client $i\in[N]$ then encodes the dataset using a Lagrange interpolation polynomial $u:\mathbb{F}_p\rightarrow \mathbb{F}_p^{\frac{m}{K}\times d}$ with degree at most $K+T-1$, 
\vspace{-0.0cm}\begin{equation}\label{eq:lag1multi} 
    [u(z)]_i 
    \triangleq \sum_{k\in[K]}[\mathbf{X}_k]_i\cdot \prod_{l\in [K+T]\setminus\{k\}}\frac{z-\beta_l}{\beta_k-\beta_l} 
    + \sum_{k=K+1}^{K+T} [\mathbf{Z}_k]_i\cdot \prod_{l\in [K+T]\setminus\{k\}}\frac{z-\beta_l}{\beta_k-\beta_l},
\end{equation} 
where $[u(\beta_k)]_i=[\mathbf{X}_k]_i$ for $k\in[K]$ and $i\in[N]$.
The matrices $\mathbf{Z}_k$ are generated uniformly at random\footnote{The random parameters can be generated by a crypto-service provider in an offline manner, or by using pseudo-random secret sharing~\cite{cramer2005share}.} from~$\mathbb{F}^{\frac{m}{K}\times d}_p$ and 
$[\mathbf{Z}_k]_i$ is the secret share of $\mathbf{Z}_k$ at client $i$.
$[\mathbf{Z}_k]_i$ is the secret share of $\mathbf{Z}_k$ at client $i$.
Client~$i\in[N]$ then computes and sends $[\widetilde{\bX}_j]_i\triangleq[u(\alpha_j)]_i$ to client $j\in[N]$. 
Upon receiving $\{[\widetilde{\bX}_j]_i\}_{i\in[N]}$,
client $j\in[N]$ can recover the encoded matrix $\widetilde{\mathbf{X}}_j$.\footnote{In fact, gathering only $T+1$ secret shares is sufficient to recover $\widetilde{\mathbf{X}}_i$, due to the construction of Shamir's secret sharing \cite{shamir1979share}. Using this fact, one can speed up the execution by dividing the $N$ clients into subgroups of $T+1$ and performing the encoding locally within each subgroup. We utilize this property in our experiments.}
The role of $\mathbf{Z}_k$'s are to mask the dataset so that the encoded matrices $\widetilde{\mathbf{X}}_j$ reveal no information about the dataset $\mathbf{X}$, even if up to $T$ clients collude, as detailed in Section \ref{sec:convergence}. 

Using the secret shares $[\mathbf{X}_j]_i$ and $[\mathbf{y}_j]_i$, clients $i\in [N]$ also compute $\bX^T \mathbf{y} = \sum_{j\in[N]}\mathbf{X}_j^T\mathbf{y}_j$ using a secure multiplication protocol (see Appendix~\ref{MPC} for details). At the end of this step, clients learn a secret share of  $\bX^T \mathbf{y}$, which we denote by $[\bX^T \mathbf{y}]_i$ for  client $i\in N$.

At iteration $t$, client $i$ initially holds a secret share of the current model, $[\mathbf{w}^{(t)}]_i$, and then encodes the model via a Lagrange interpolation polynomial $v:\mathbb{F}_p\rightarrow \mathbb{F}_p^{d}$ with degree at most $K+T-1$,
\begin{equation}\label{eq:lag2multi}
    [v(z)]_i 
    \triangleq \sum_{k\in[K]}[\mathbf{w}^{(t)}]_i\cdot \prod_{l\in [K+T]\setminus\{k\}}\frac{z-\beta_l}{\beta_k-\beta_l} 
    +\sum_{k=K+1}^{K+T} [\mathbf{v}^{(t)}_k]_i\cdot \prod_{l\in [K+T]\setminus\{k\}}\frac{z-\beta_l}{\beta_k-\beta_l},
\end{equation} 
where $[v(\beta_{k})]_i=[\mathbf{w}^{(t)}]_i$ for $k\in [K]$ and $i\in[N]$. 
The vectors $\mathbf{v}^{(t)}_k$ are generated uniformly at random from~$\mathbb{F}^{d}_p$.
Client~$i\in[N]$ then sends   
$[\widetilde{\mathbf{w}}^{(t)}_j]_i\!\triangleq\![v(\alpha_j)]_i$ to client~$j\in[N]$.
Upon receiving  $\{[\widetilde{\mathbf{w}}^{(t)}_j]_i\}_{i\in[N]}$, client $j\in[N]$ recovers the encoded model $\widetilde{\mathbf{w}}^{(t)}_j$. 


\vspace{0.1cm}
\noindent
{\bf Phase 3: Polynomial Approximation and Local  Computations.} 
Lagrange encoding can be used to compute polynomial functions only, whereas the gradient computations in \eqref{grad} are not polynomial operations due to the sigmoid function. To this end, we approximate the sigmoid with a polynomial,
\begin{equation}\label{eq:poly_approximation} 
\hat{g}(z) = \sum_{i=0}^{r} c_{i} z^{i},
\end{equation}
where $r$ and $c_i$ represent the degree and coefficients of the polynomial, respectively. The coefficients are evaluated by fitting the sigmoid to the polynomial function via least squares estimation. 
Using this polynomial approximation, we rewrite the model update from \eqref{grad} as,
\begin{equation}\label{eq:apprx_grad1multi} 
\bw^{(t+1)}  =   \bw^{(t)}-\frac{\eta}{m}\mathbf{X}^{\top}(\hat{g}(\mathbf{X} \times \mathbf{w}^{(t)}) -\mathbf{y}).
\end{equation} 
Client $i\in [N]$ then locally computes the gradient over the encoded dataset, by evaluating a function,
\begin{equation}\label{eq:local}
f(\widetilde{\bX}_i, \widetilde{\mathbf{w}}^{(t)}_i)=\widetilde{\bX}_i^\top \hat{g}(\widetilde{\bX}_i\times \widetilde{\mathbf{w}}^{(t)}_i) 
\end{equation}
and secret shares the  result with the other clients, by sending a secret share of \eqref{eq:local}, $[f(\widetilde{\bX}_i, \widetilde{\mathbf{w}}^{(t)}_i)]_j$, to client $j\in [N]$. 
At the end of this step, client $j$ holds the secret shares $[f(\widetilde{\bX}_i, \widetilde{\mathbf{w}}^{(t)}_i)]_j$ corresponding to the local computations from clients $i\in [N]$. 
Note that \eqref{eq:local} is a polynomial function evaluation in the finite field arithmetic and the degree of function $f$ is $\text{deg}(f) = 2r+1$.

\vspace{0.1cm}
\noindent
{\bf Phase 4: Decoding and Model Update.}  
In this phase, clients perform the decoding of the gradient using a secure MPC protocol, through polynomial interpolation over the secret shares $[f(\widetilde{\bX}_i, \widetilde{\mathbf{w}}^{(t)}_i)]_j$.
The minimum number of clients needed for the decoding operation to be successful, which we call the \emph{recovery threshold} of the protocol, is equal to $(2r+1)(K+T-1)+1$. 
In order to show this, we first note that, from the definition of Lagrange polynomials in \eqref{eq:lag1multi}  and \eqref{eq:lag2multi}, one can define
a univariate polynomial $h(z)=f\big(u(z),v(z)\big)$ such that
\begin{equation}
    h(\beta_i)=f\big(u(\beta_i),v(\beta_i)\big) 
    =f\big({\bX}_i,{\bw}^{(t)}\big) 
    = {\mathbf{X}}_i^{\top}\hat{g}({\mathbf{X}}_i\times {\mathbf{w}}^{(t)}) \label{eq:beta}
\end{equation}
for $i\in[K]$. Moreover, from \eqref{eq:local}, we know that client $i$ performs the following computation, 
\begin{equation}
    h(\alpha_i)=f\big(u(\alpha_i),v(\alpha_i)\big)
    =  f\big(\widetilde{\bX}_i,\widetilde{\bw}^{(t)}_i\big) 
    =\widetilde{\bX}^{\top}_i \hat{g}(\widetilde{\bX}_i \times  \widetilde{\bw}^{(t)}_i). 
\label{eq:alpha}
\end{equation} 
The decoding process is based on the intuition that, the computations from \eqref{eq:alpha} can be used as evaluation points $h(\alpha_i)$ to interpolate the polynomial $h(z)$. Since the degree of the polynomial $h(z)$ is $\text{deg}\big( h(z) \big) \leq (2r+1)(K+T-1)$, all of its coefficients can be determined as long as there are at least $(2r+1)(K+T-1)+1$ evaluation points available. 
After $h(z)$ is recovered, the computation results in \eqref{eq:beta} correspond to $h(\beta_i)$ for  $i\in[K]$. 

Our decoding operation corresponds to a finite-field polynomial interpolation problem. 
More specifically, upon receiving the secret shares of the local computations $[f(\widetilde{\bX}_j, \widetilde{\mathbf{w}}^{(t)}_j)]_i$ from at least $(2r+1)(K+T-1)+1$  clients, client $i$ locally computes
\begin{equation}\label{eq:local_dec}
    [f(\bX_k, \bw^{(t)})]_i \triangleq \sum_{j\in\mathcal{I}_i}[f(\widetilde{\bX}_j, \widetilde{\mathbf{w}}^{(t)}_j)]_i\cdot \prod_{l\in \mathcal{I}_i\setminus\{j\}}\frac{\beta_k-\alpha_l}{\alpha_j-\alpha_l}
\end{equation}
for $k\in[K]$, where $\mathcal{I}_i\subseteq[N]$ denotes the set of the $(2r+1)(K+T-1)+1$ fastest clients who  send their secret share  $[f(\widetilde{\bX}_j, \widetilde{\mathbf{w}}^{(t)}_j)]_i$ to client $i$. 



After this step, client $i$ locally aggregates its secret shares $[f(\bX_k, \bw^{(t)})]_i$ to compute  $\sum_{k=1}^K[f(\bX_k, \bw^{(t)})]_i$,   
which in turn is a secret share of $\bX^T \hat{g}(\bX \times \bw^{(t)})$ since,
\begin{equation}
    \sum_{k=1}^K f(\bX_k, \bw^{(t)}) = \sum_{k=1}^K \bX_k^\top \hat{g}(\bX_k \times \bw^{(t)}) 
    = \bX^\top \hat{g}(\bX \times \bw^{(t)}).  
    \label{eq:agg}
\end{equation}
Let $[\bX^\top\hat{g}(\bX\times \bw^{(t)})]_i \triangleq \sum_{k=1}^K[f(\bX_k, \bw^{(t)})]_i$ denote the secret share of \eqref{eq:agg} at client $i$.   
Client $i$ then computes  $[\bX^\top\hat{g}(\bX\times \bw^{(t)})]_i - [\bX^\top \mathbf{y}]_i$, which in turn is a secret share of the gradient  $\mathbf{X}^{\top}(\hat{g}(\mathbf{X} \times \mathbf{w}^{(t)}) -\mathbf{y})$. 
Since the decoding operations are carried out using the secret shares, at the end of the decoding process, the clients only learn a secret share of the gradient and not its true value.

Next, clients update the model according to \eqref{eq:apprx_grad1multi} using a secure MPC protocol, using the secret shared model $[\bw^{(t)}]_i$ and the secret share of the gradient $[\bX^\top\hat{g}(\bX\times \bw^{(t)})]_i - [\bX^\top \mathbf{y}]_i$. 
A major challenge in performing the  model update in  \eqref{eq:apprx_grad1multi}  in the finite field is the multiplication with parameter $\frac{\eta}{m}$, where $\frac{\eta}{m} < 1$. In order to perform this operation in the finite field, one potential approach is to treat it as a computation on integer numbers and preserve full accuracy of the results. 
This in turn requires a very large field size as the range of results grows exponentially with the number of multiplications, which becomes quickly impractical as the number of iterations increase \cite{mohassel2017secureml}. Instead, we address this problem by leveraging the secure truncation technique from \cite{catrina2010secure}. 
This protocol takes secret shares $[a]_i$ of a variable $a$ as input as well as two public integer parameters $k_1$ and $k_2$ such that  
$a \in \mathbb{F}_{2^{k_2}}$ and $0< k_1 < k_2$. The protocol then returns the secret shares $[z]_i$ for $i\in [N]$  such that $z = \lfloor\frac{a}{2^{k_1}}\rfloor + s$  where $s$ is a random bit with probability $P(s=1) = (a \mod 2^{k_1})/(2^{k_1})$. Accordingly, the protocol rounds $a/(2^{k_1})$ to the closest integer with probability $1-\tau$, with $\tau$ being the distance between $a/(2^{k_1})$ and that integer. 
The truncation operation ensures that the range of the updated model always stays within the range of the finite field.

Since the model update is carried out using a secure MPC protocol, at the end of this step, client $i\in [N]$ learns only a secret share $[\bw^{(t+1)}]_i$ of the updated model $\bw^{(t+1)}$, and not its actual value. 
In the next iteration, using $[\bw^{(t+1)}]_i$, client $i\in[N]$ locally computes $[\widetilde{\bw}^{(t+1)}_j]_i$ from \eqref{eq:lag2multi} and sends it to client $j\in[N]$. Client $j$ then recovers the encoded model  $\widetilde{\bw}^{(t+1)}_j$, which is used to compute \eqref{eq:local}. 

The implementation details of the MPC protocols are provided in Appendix~\ref{MPC}. 
The overall algorithm for {\acr} is presented in Appendix~\ref{app:algorithms}. 


\section{Convergence and Privacy Guarantees}\label{sec:convergence}

\vspace{-0.1cm}
Consider the cost function in \eqref{cost} with the quantized dataset, and denote $\bw^{*}$ as the optimal model parameters that minimize~\eqref{cost}. 
In this subsection, we prove that {\acr} guarantees convergence to the optimal model parameters (i.e., $\bw^{*}$) while maintaining the privacy of the dataset against colluding clients. This result is stated in the following theorem.
\begin{theorem}\label{thm:convergence} For training a logistic regression model in a distributed system with $N$ clients using the quantized dataset $\mathbf{X} = [\mathbf{X}_{1}^\top, \ldots, \mathbf{X}_{N}^\top]^\top$, initial model parameters $\bw^{(0)}$,  and constant step size $\eta \leq 1/L$ (where $L=\frac{1}{4}\| {\bX} \|^{2}_{2}$), {\acr} guarantees convergence,
\vspace{-0.1cm}
\begin{equation} \mathbb{E}\big[ C\big( \frac{1}{J}\sum_{t=0}^{J}\bw^{(t)}\big)\big]-C(\bw^{*})\leq 
    \frac{{\| \bw^{(0)}-\bw^{*} \|}^2}{2\eta J}+\eta \sigma^2 
    \end{equation}
in $J$ iterations, for any $N\geq (2r+1)(K+T-1) +1$, where $r$ is the degree of the polynomial in~\eqref{eq:poly_approximation} and $\sigma^2$ is the variance of the quantization error of the secure truncation protocol.
\end{theorem}

\vspace{-0.3cm}\begin{proof}\normalfont
The proof of Theorem~\ref{thm:convergence} is presented in Appendix~\ref{app:convergenceproof}.
\end{proof}\vspace{-0.3cm}

As for the privacy guarantees, {\acr} protects the statistical privacy of the individual  dataset of each client against up to $T$ colluding adversarial clients, even if the adversaries have unbounded computational power. 
The privacy protection of {\acr} follows from the fact that all building blocks of the algorithm guarantees either (strong) information-theoretic privacy or statistical privacy of the individual datasets against any collusions between up to $T$ clients.
Information-theoretic privacy of Lagrange coding against $T$ colluding clients follows from \cite{yu2018lagrange}. Moreover, encoding, decoding, and model update operations are carried out in a secure MPC protocol that protects the information-theoretic privacy of the corresponding computations against $T$ colluding clients \cite{ben1988completeness,beerliova2008perfectly,damgaard2007scalable}. Finally, the (statistical) privacy guarantees of the truncation protocol follows from \cite{catrina2010secure}.  


\begin{remark}\normalfont
(Privacy-parallelization trade-off)
Theorem~\ref{thm:convergence} reveals an important trade-off between privacy and parallelization in {\acr}.
Parameter $K$ reflects the amount of parallelization. 
In particular, the size of the encoded matrix at each client is equal to $(1/K)^{th}$ of the size of $\bX$. Since each client computes the gradient over the encoded dataset, the computation load at each client is proportional to processing $(1/K)^{th}$ of the entire dataset. As $K$ increases, the computation load at each client decreases.
Parameter $T$ reflects the privacy threshold of {\acr}. In a distributed system with $N$ clients, {\acr} can achieve any  $K$ and $T$ as long as $N\geq (2r+1)(K+T-1) +1$. Moreover, as the number of clients $N$ increases, parallelization ($K$) and privacy ($T$) thresholds of {\acr} can also increase linearly, providing a scalable solution. The motivation behind the encoding process is to distribute the load of the computationally-intensive gradient evaluations across multiple clients (enabling parallelization), and to protect the privacy of the dataset.
\end{remark} 

\begin{remark}\normalfont 
Theorem~\ref{thm:convergence} also holds for the simpler linear regression problem.
\end{remark} 

\section{Experiments}\label{sec:experiments} 

\begin{figure}%
    \centering
    \subfigure[CIFAR-10 (for accuracy $80.45\%$)]{%
    \label{fig:Masterless_first}%
    \includegraphics[height=1.8in, trim=0.2cm 0.2cm 0.2cm 0.2cm, clip]{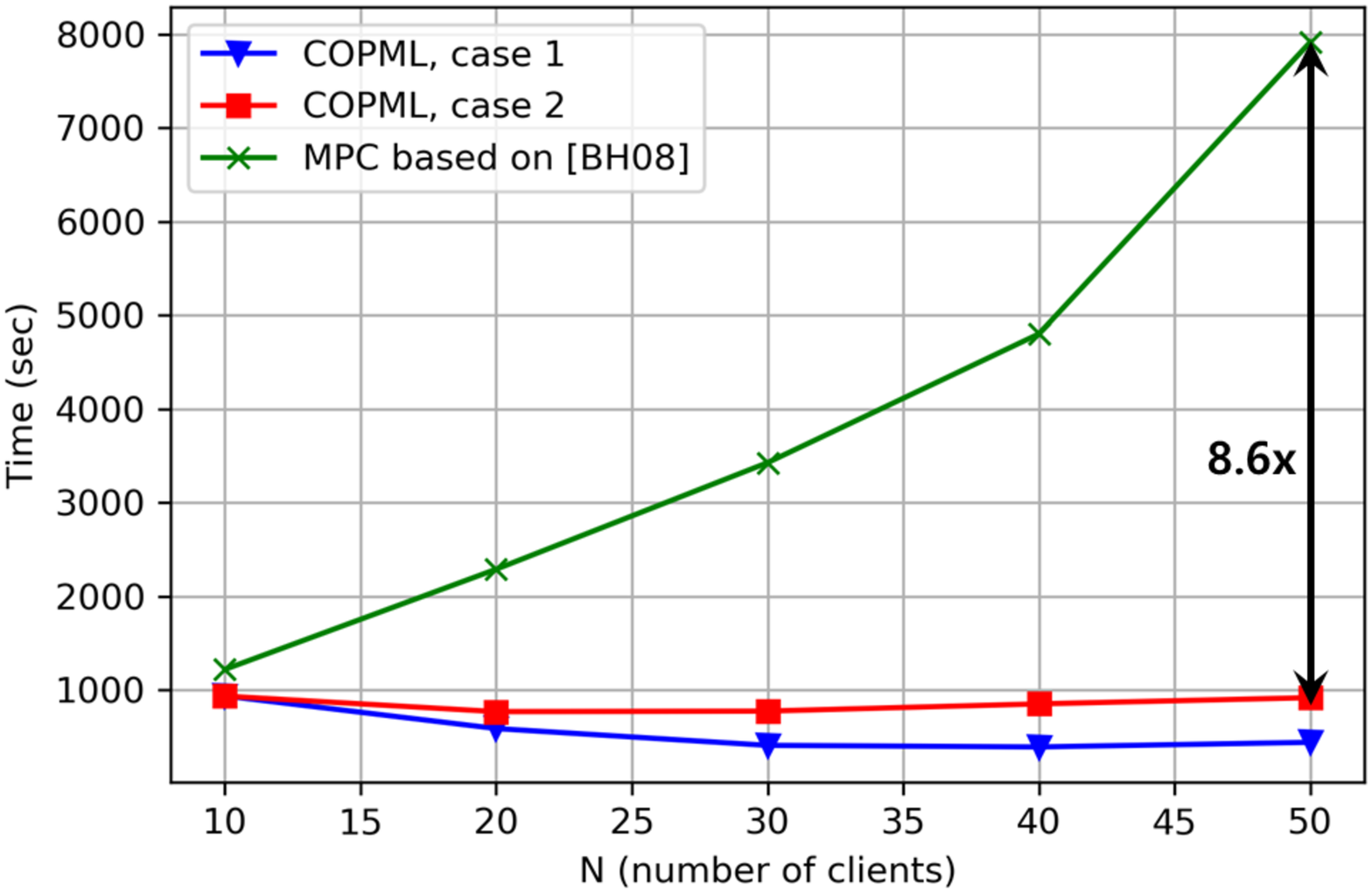}}%
    \;\;
    \subfigure[GISETTE (for accuracy $97.50\%$)]{%
    \vspace{-0.3cm}
    \label{fig:Masterless_second}%
    \includegraphics[height=1.8in, trim=0.2cm 0.2cm 0.2cm 0.2cm, clip]{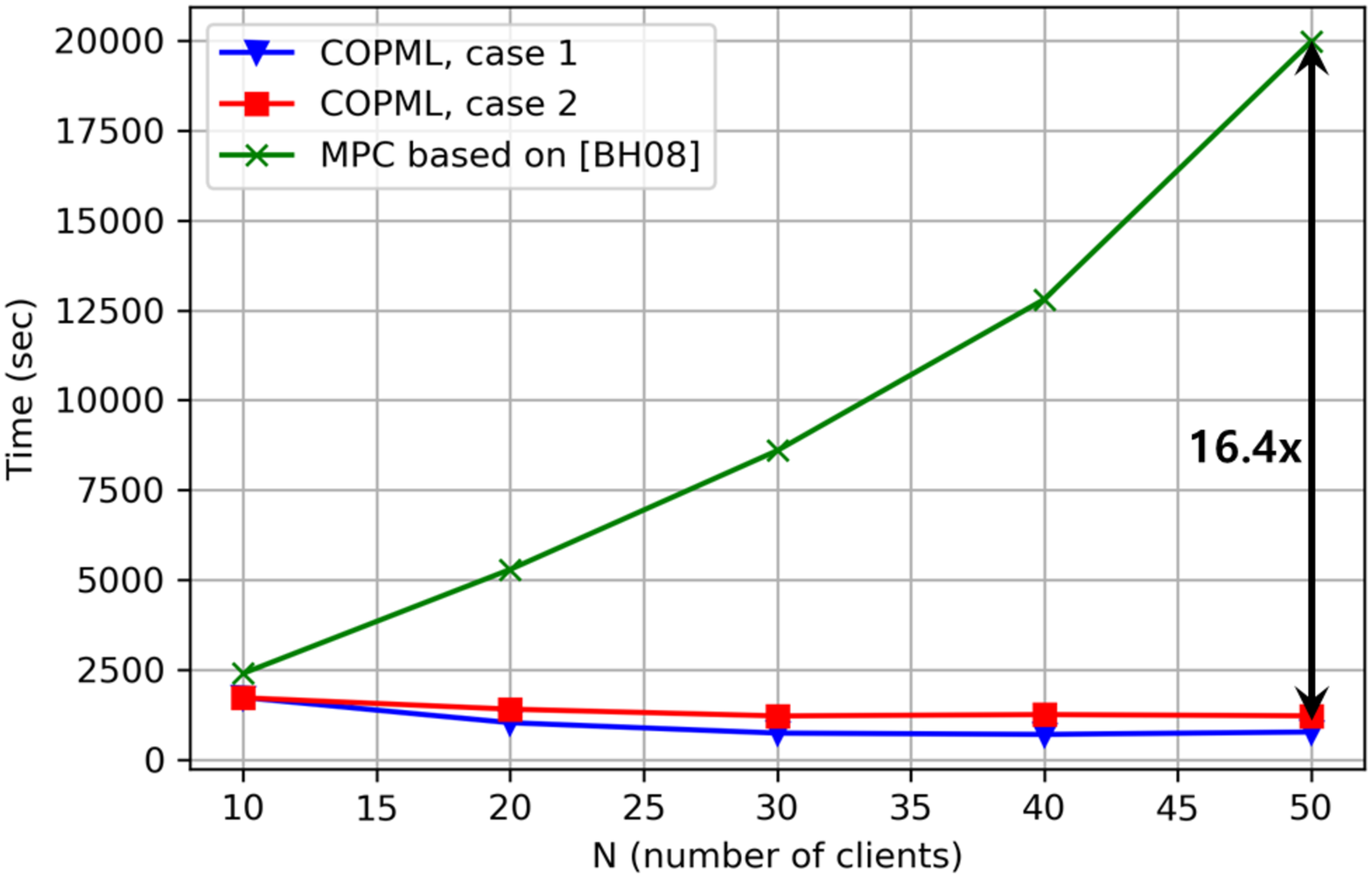}}%
    \vspace{-0.2cm}
    \caption{Performance gain of {\acr} over the MPC baseline ([BH08] from \cite{beerliova2008perfectly}). The plot shows the total training time for different number of clients $N$  with $50$ iterations. 
}
\label{fig:trainingtimeSec5}
\vspace{-0.4cm}
\end{figure}

We demonstrate the performance of {\acr} compared to conventional MPC baselines by examining two properties, accuracy and performance gain, in terms of the  training time on the Amazon EC2 Cloud Platform.
\vspace{-0.2cm}
\subsection{Experiment setup}
\vspace{-0.1cm}
{\bf Setup.}
We train a logistic regression model for binary image classification on the CIFAR-10 \cite{krizhevsky2009learning} and GISETTE~\cite{NIPS2004_2728} datasets, whose size is $(m,d)=(9019,3073)$ and $(6000,5000)$, respectively.
The dataset is distributed evenly across the clients. The clients initially secret share their individual datasets with the other  clients.\footnote{This can be done offline as it is an identical one-time operation for both MPC baselines and {\acr}.}
Computations are carried out on Amazon EC2 \texttt{m3.xlarge} machine instances.
We run the experiments in a WAN setting with an average bandwidth of $40Mbps$.
Communication between clients is implemented using the {\tt MPI4Py}~\cite{dalcin2011parallel} interface on {\tt Python}.

{\color{black}
{\bf Implemented schemes.} We implement four schemes for performance evaluation. For {\acr}, we consider two set of key parameters $(K,T)$ to investigate the trade-off between parallelization and privacy. For the baselines, we apply two conventional MPC protocols (based on \cite{ben1988completeness} and  \cite{beerliova2008perfectly}) to our multi-client problem setting.\footnote{As described in the Section \ref{sec:intro}, there is no prior work at our scale (beyond 3-4 parties), hence we implement two baselines based on well-known MPC protocols which are also the first implementations at our scale.}
\vspace{-0.1cm}
\begin{enumerate}[leftmargin=0.6cm]
    \item {\bf {\acr}.} 
    In {\acr}, MPC is utilized to enable secure encoding and decoding for Lagrange coding.
    The gradient computations are then carried out using the Lagrange encoded data. 
    We determine $T$ (privacy threshold) and $K$ (amount of parallelization) in {\acr} as follows. Initially, we have from Theorem~\ref{thm:convergence} that  these parameters must satisfy $N\geq (2r+1)(K+T-1) +1$ for our framework. 
    Next, we have considered both $r=1$ and $r=3$ for the  degree of the polynomial approximation of the sigmoid function and observed that the degree one approximation achieves good accuracy, as we demonstrate later.
    Given our choice of $r=1$, we then consider two setups: 
    
\noindent\textbf{Case 1:} \emph{(Maximum parallelization gain)} 
    Allocate all resources to parallelization (fastest  training), by letting $K= \lfloor \frac{N\!-\!1}{3} \rfloor$ and $T=1$,
    
\noindent
    \textbf{Case 2:} \emph{(Equal parallelization and privacy gain)} Split resources almost equally between parallelization and privacy, i.e.,
     $T=\lfloor \frac{N-3}{6} \rfloor, K=\lfloor \frac{N+2}{3} \rfloor-T$. 
    
\item {\bf Baseline protocols.}
    We implement two conventional MPC protocols (based on \cite{ben1988completeness} and  \cite{beerliova2008perfectly}). 
    In a naive implementation of these protocols, each client would secret share its local dataset with the entire set of clients, and the gradient computations would be performed over the secret shared data whose size is as large as the entire dataset, which leads to a significant computational overhead. 
    For a fair comparison with {\acr}, we speed up the baseline protocols by partitioning the clients into three groups, and assigning each group one third of the entire dataset.
    Hence, the total amount of data processed at each client is equal to one third of the size of the entire dataset, which significantly reduces the total training time while providing a privacy threshold of $T=\lfloor \frac{N-3}{6} \rfloor$, which is the same privacy threshold as Case 2 of {\acr}.
    The details of these implementations are presented in Appendix~\ref{app:opt_baseline}.

\vspace{-0.1cm}
\end{enumerate}

In all schemes, we apply the MPC truncation protocol from Section~\ref{app:multiclient_framework} to carry out the multiplication with  $\frac{\eta}{m}$ during model updates, by choosing $(k_1,k_2)=(21, 24)$ and $(22, 24)$ for the CIFAR-10 and GISETTE datasets, respectively. 
}



\begin{figure}[t!]%
    \centering
    \subfigure[CIFAR-10 dataset for binary classification between \emph{plain} and \emph{car} images (using $9019$ samples for the training set and $2000$ samples for the test set).]{%
    \vspace{-0.7cm}
    \label{fig:accuracy_masterless_CIFAR}%
    \includegraphics[height=1.4in, trim=0.2cm 0.2cm 0.3cm 0.3cm, clip]{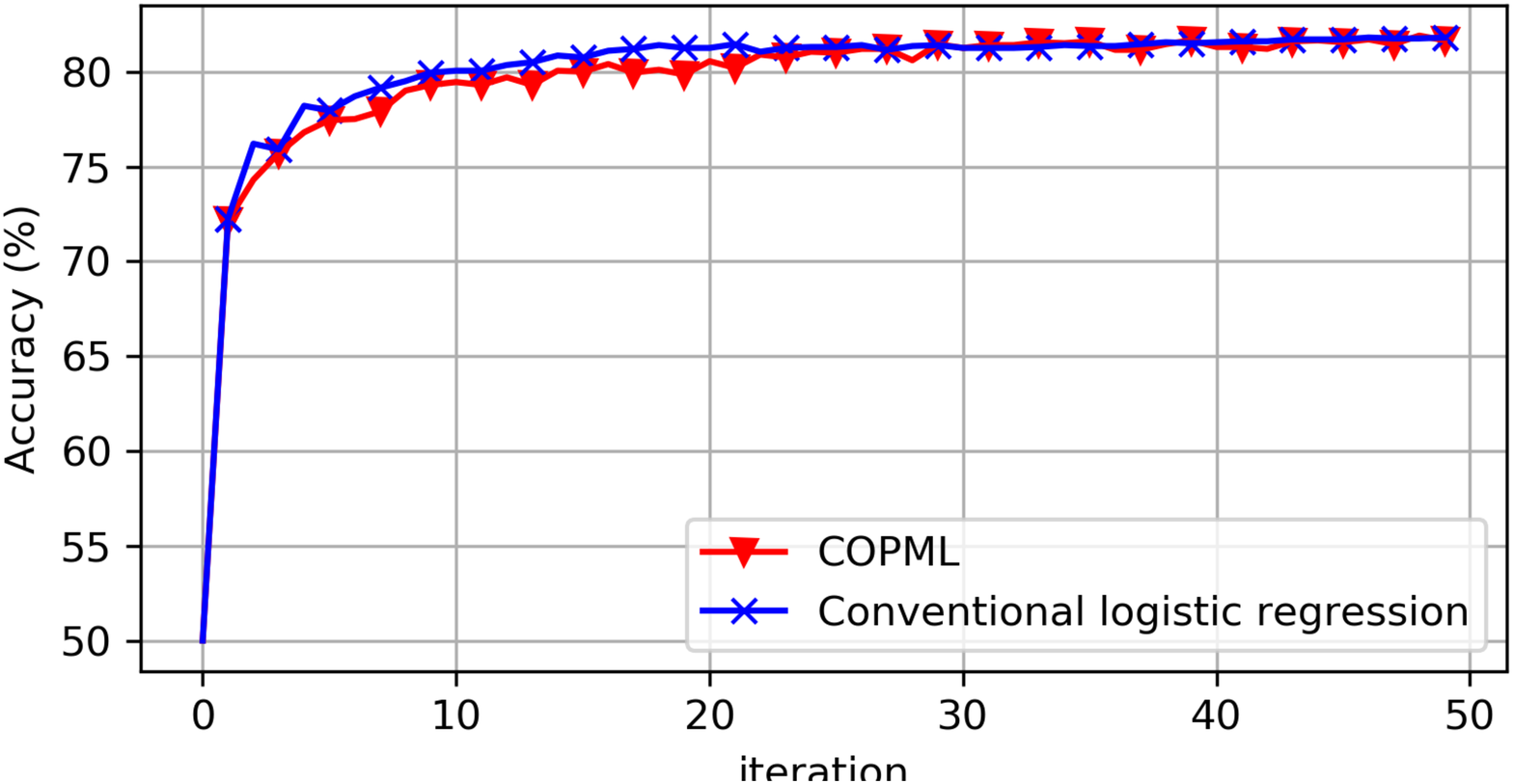}}%
    \;\;
    \subfigure[GISETTE dataset for binary classification between digits $4$ and $9$ (using $6000$ samples for the training set and $1000$ samples for the test set).]{%
    \vspace{-0.7cm}
    \label{fig:accuracy_masterless_GISETTE}%
    \includegraphics[height=1.4in, trim=0.2cm 0.2cm 0.3cm 0.3cm, clip]{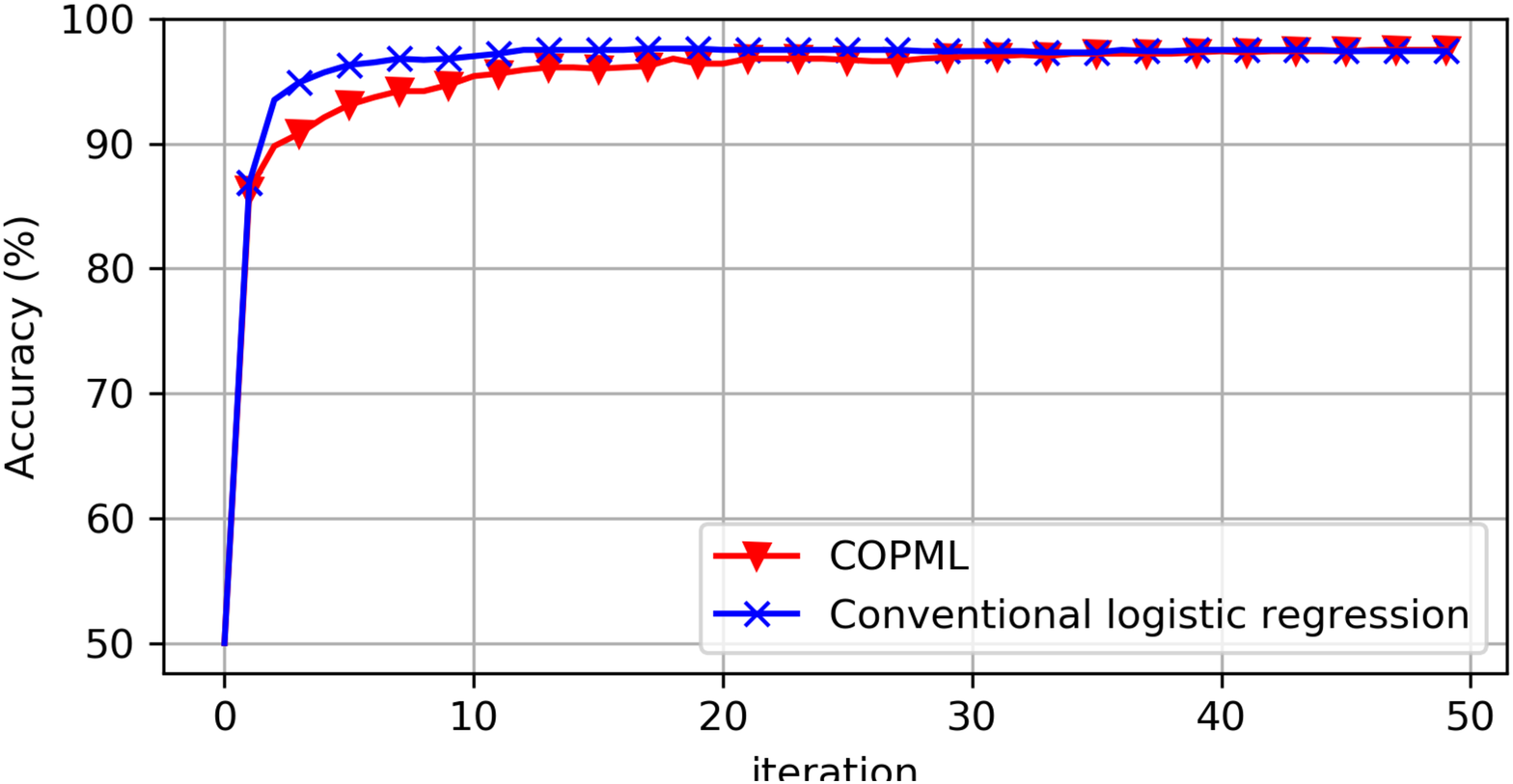}}%
    \vspace{-0.2cm}
\caption{Comparison of the accuracy of {\acr} (demonstrated for Case 2 and $N=50$ clients) vs  conventional logistic regression that uses the sigmoid function without quantization.}
\label{fig:accuracySec5}
\vspace{-0.4cm}
\end{figure}

\subsection{Performance evaluation}

\begin{table}[t!]
    \footnotesize
    \caption{Breakdown of the running time with $N=50$ clients.}
    \vspace{0.1cm}
    \label{table:N50}
    \centering
    \begin{tabular}{ c | c | c | c | c }
    \toprule
    \multirow{2}{*}{Protocol} 
     & Comp. & Comm. &  Enc/Dec & Total run \\
     & time (s)& time (s) & time (s)& time (s)\\
     \midrule
    MPC using [BGW88]& \hspace{-0.1cm} 918 & \hspace{-0.0cm}21142 & 324 & 22384 \\ 
    MPC using [BH08] & \hspace{-0.1cm} 914 & \hspace{-0.0cm}6812 & 189 & 7915 \\ 
    {\acr} (Case 1) &  \hspace{-0.1cm} 141 & \hspace{0.0cm}284 &  15 & 440  \\ 
    {\acr} (Case 2) &  \hspace{-0.1cm} 240 & \hspace{0.0cm}654 &  22 & 916  \\ 
    \bottomrule
    \end{tabular}
\end{table}

{\bf Training time.}
In the first set of experiments, we measure the training time. 
Our results are demonstrated in Figure~\ref{fig:trainingtimeSec5}, which shows the comparison of {\acr} with the protocol from~\cite{beerliova2008perfectly}, as we have found it
to be the faster of the two baselines.
Figures~\ref{fig:Masterless_first} and \ref{fig:Masterless_second} demonstrate that {\acr} provides substantial speedup over the MPC baseline, in particular, up to $8.6\times$ and $16.4\times$ with the CIFAR-10 and GISETTE datasets, respectively, while providing the same privacy threshold $T$.
We observe that a higher amount of speedup is achieved as the dimension
of the dataset becomes larger (CIFAR-10 vs. GISETTE datasets),
suggesting {\acr} to be well-suited for data-intensive distributed training tasks where parallelization is essential.

To further investigate the gain of {\acr}, in Table~\ref{table:N50} we present the breakdown of the total running time with the  CIFAR-10 dataset for $N=50$ clients. We observe that {\acr} provides $K/3$ times speedup for the computation time of matrix multiplication in~\eqref{eq:local}, which is given in the first column. This is due to the fact that, in the baseline protocols, the size of the data processed at each client is one third of the entire dataset, while in {\acr} it is $(1/K)^{th}$ of the entire dataset. This reduces the computational overhead of each client while computing matrix multiplications. Moreover, {\acr} provides significant improvement in the  communication, encoding, and decoding time. This is because the two baseline protocols require intensive communication and computation 
to carry out a degree reduction step for secure multiplication (encoding and decoding for additional secret shares), which is detailed in Appendix~\ref{MPC}. 
In contrast, {\acr} only requires secure addition and multiplication-by-a-constant operations for encoding and decoding. 
These operations require no communication.
In addition, the communication, encoding, and decoding overheads of each client are also reduced due to the fact that the size of the data processed at each client is only $(1/K)^{th}$ of the entire dataset. 



{\bf Accuracy.} We finally examine the accuracy of {\acr}. Figures~\ref{fig:accuracy_masterless_CIFAR} and \ref{fig:accuracy_masterless_GISETTE} demonstrate that {\acr} with degree one polynomial approximation provides comparable test accuracy to conventional logistic regression. For the CIFAR-10 dataset in Figure~\ref{fig:accuracy_masterless_CIFAR}, the accuracy of {\acr} and conventional logistic regression are $80.45\%$ and $81.75\%$, respectively, in $50$ iterations. For the GISETTE dataset in Figure~\ref{fig:accuracy_masterless_GISETTE}, the accuracy of {\acr} and conventional logistic regression have the same value of $97.5\%$ in 50 iterations.
Hence, {\acr} has comparable accuracy to conventional logistic regression while also being privacy preserving.

\begin{table}[t!]
\small
\caption{Complexity summary of {\acr}. }\label{tbl:complexity}
\vspace{-0.3cm}
\begin{center}
\begin{tabular}{ccc}
\toprule
  Communication & Computation & Encoding   \\
\midrule
 $O(\frac{mdN}{K} + dNJ)$  & $O(\frac{md^2}{K})$ & $O(\frac{mdN(K+T)}{K}+dN(K+T)J)$ \\
\bottomrule
\end{tabular}
\end{center}
\vspace{-0.5cm}
\end{table}

{
\subsection{Complexity Analysis}\label{sec:complexity}
\vspace{-0.2cm}
In this section, we analyze the asymptotic complexity of each client in {\acr} with respect to the number of users $N$, model dimension $d$, number of data points $m$, parallelization parameter $K$, privacy parameter $T$, and total number of iterations $J$.
Client $i$'s communication cost can be broken to three parts: 1) sending the secret shares $[\widetilde{\mathbf{X}}_j]_i=[u(\alpha_j)]_i$ in \eqref{eq:lag1multi} to client $j\in[N]$, 2) sending the secret shares $[\widetilde{\mathbf{w}}^{(t)}_j]_i=[v(\alpha_j)]_i$ in \eqref{eq:lag2multi} to client $j\in[N]$ for $t\in \{0,\ldots,J-1\}$, and 3) sending the secret share of local computation $[f(\widetilde{\mathbf{X}}_i,\widetilde{\mathbf{w}}^{(t)}_i)]_j$ in \eqref{eq:local} to client $j\in[N]$ for $t\in \{0,\ldots,J-1\}$.
The communication cost of the three parts are $O(\frac{mdN}{K})$, $O(dNJ)$, and $O(dNJ)$, respectively. Therefore, the overall communication cost of each client is $O(\frac{mdN}{K} + dNJ)$.
User $i$'s computation cost of encoding can be broken into two parts, encoding the dataset by using \eqref{eq:lag1multi} and encoding the model by using \eqref{eq:lag2multi}.
The encoded dataset $[\widetilde{\mathbf{X}}_j]_i=[u(\alpha_j)]_i$ from \eqref{eq:lag1multi} is a weighted sum of $K+T$ matrices where each matrix belongs to $\mathbb{F}_p^{\frac{m}{K}\times d}$. As there are $N$ encoded dataset and each encoded dataset requires a computation cost of $O(\frac{md(K+T)}{K})$, the computation cost of encoding the dataset is $O(\frac{mdN(K+T)}{K})$ in total. Similarly, computation cost of encoding $[\widetilde{\mathbf{w}}^{(t)}_j]_i=[v(\alpha_j)]_i$ from \eqref{eq:lag2multi} is $O(dN(K+T)J)$.
Computation cost of client $i$ to compute $\widetilde{\mathbf{X}}_i^{\top}\widetilde{\mathbf{X}}_i$, the dominant part of local computation $f(\widetilde{\mathbf{X}}_i,\widetilde{\mathbf{w}}^{(t)}_i)$ in \eqref{eq:local}, is $O(\frac{md^2}{K})$.
We summarize the asymptotic complexity of each client in Table~\ref{tbl:complexity}.
}

{
When we set $N=3(K+T-1)+1$ and $K=O(N)$ (Case 2), increasing $N$ has two major impacts on the training time: 1) reducing the computation per worker by choosing a larger $K$, 2) increasing the encoding time.
In this case, as $m$ is typically much larger than other parameters, dominate terms in communication, computation, and encoding cost are $O(md)$, $O(md^2/N)$ and $O(mdN)$, respectively. 
For small datasets, i.e., when the computation load at each worker is very small, the gain from increasing the number of workers beyond a certain point may be minimal and system may saturate, as encoding may dominate the computation. 
This is the reason that a higher amount of speedup of training time is achieved as the dimension of the dataset becomes larger.
}
\vspace{-0.15cm}


\section{Conclusions} 
\vspace{-0.3cm}
{
We considered a collaborative learning scenario in which multiple data-owners jointly train a logistic regression model without revealing their individual datasets to the other parties.
To the best of our knowledge, even for the simple logistic regression, {\acr} is the first fully-decentralized training framework to scale beyond 3-4 parties while achieving information-theoretic privacy.
Extending {\acr} to more complicated (deeper) models is a very interesting future direction.
An MPC-friendly (i.e., polynomial) activation function is proposed in \cite{mohassel2017secureml} which approximates the softmax and shows that the accuracy of the resulting models is very close to those trained using the original functions. 
We expect to achieve a similar performance gain even in those setups, since {\acr} can similarly be leveraged to efficiently parallelize the MPC computations. }

\clearpage

\section*{Broader Impact}
Our framework has the societal benefit of protecting user privacy in collaborative machine learning applications, where multiple data-owners can jointly train machine learning models without revealing information about their individual datasets to the other parties, even if some parties collude with each other. Collaboration can significantly improve the accuracy of trained machine learning models, compared to training over individual datasets only. This is especially important in applications where data labelling is costly and can take a long time, such as data collected and labeled in medical fields. For instance, by using our framework, multiple medical institutions can collaborate to train a logistic regression model jointly, without revealing the privacy of their datasets to the other parties, which may contain sensitive patient healthcare records or genetic information. Our framework can scale to a significantly larger number of users compared to the benchmark protocols, and can be applied to any field in which the datasets contain sensitive information, such as healthcare records, financial transactions, or geolocation data. In such applications, protecting the privacy of sensitive information is critical and failure to do so can result in serious societal, ethical, and legal consequences. Our framework can provide both application developers and users with positive societal consequences, application developers can provide better user experience with better models as the volume and diversity of data will be increased greatly, and at the same time, users will have their sensitive information kept private. Another benefit of our framework is that it provides strong privacy guarantees that is independent from the computational power of the adversaries. Therefore, our framework keeps the sensitive user information safe even if adversaries gain quantum computing capabilities in the future.  

A potential limitation of our framework is that our current training framework is bound to polynomial operations. In order to compute functions that are not polynomials, such as the sigmoid function, we utilize a polynomial approximation. This can pose a challenge in the future for applying our framework to deep neural network models, as the approximation error may add up at each layer. In such scenarios, one may need to develop additional techniques to better handle the non-linearities and approximation errors.


\section*{Acknowledgement}
This material is based upon work supported by Defense Advanced Research Projects Agency (DARPA) under Contract No. HR001117C0053, ARO award W911NF1810400, NSF grants CCF-1703575 and CCF-1763673, ONR Award No. N00014-16-1-2189, and research gifts from Intel and Facebook. The views, opinions, and/or findings expressed are those of the author(s) and should not be interpreted as representing the official views or policies of the Department of Defense or the U.S. Government.

\clearpage

\bibliographystyle{unsrt} 
\bibliography{main}

\clearpage
\appendix
\section{Supplementary Materials}

\subsection{Details of the Quantization Phase}~\label{app:quantization}
\hspace{-0.25cm}For quantizing its dataset $\mathbf{X}_j$, client $j\in[N]$ employs a scalar quantization function $\phi\left(Round(2^{l_x} \cdot \mathbf{X}_j)\right)$, where the rounding operation
\vspace{-0.0cm}\begin{equation}\label{eq:round} 
    Round(x) =
    \left\{
    \begin{array}{ll}
          \lfloor x \rfloor & \text{if \quad } x-\lfloor x \rfloor < 0.5 \\
          \lfloor x \rfloor+1 & \text{otherwise } 
    \end{array} 
    \right. 
\vspace{-0.0cm}\end{equation}
is applied element-wise to the elements $x$ of matrix $\bX_j$ and $l_x$ is an integer parameter to control the quantization loss. 
$\lfloor x \rfloor$ is the largest integer less than or equal to $x$, and function $\phi:\mathbb{Z}\rightarrow\mathbb{F}_{p}$ is a mapping defined to represent a negative integer in the finite field by using two's complement representation,
\vspace{-0.0cm}\begin{equation}\label{eq:phi} 
    \phi(x) =
    \left\{
    \begin{array}{ll}
          x & \text{if } x \geq 0\\
          p+x & \text{if } x<0
    \end{array} 
    \right. 
\end{equation}
To avoid a wrap-around which may lead to an overflow error, prime $p$ should be large enough, $p\geq 2^{l_x+1}\max \{ \lvert x \rvert \} +1$.
Its value also depends on the bitwidth of the machine as well as the dimension of the dataset. For example, in a $64$-bit implementation with the CIFAR-10 dataset whose dimension is $d=3072$, we select $p=2^{26}-5$, which is the largest prime needed to avoid an overflow on intermediate multiplications. In particular, in order to speed up the running time of matrix-matrix multiplication, we do a modular operation after the inner product of vectors instead of doing a modular operation per product of each element. To avoid an overflow on this, $p$ should be smaller than a threshold given by $d(p-1)^2 \leq 2^{64} -1$. For ease of exposition, throughout the paper, $\mathbf{X} = [\mathbf{X}_{1}^\top, \ldots, \mathbf{X}_{N}^\top]^\top$ refers to the quantized dataset. 

\subsection{Proof of Theorem~\ref{thm:convergence}}~\label{app:convergenceproof}
\hspace{-0.11cm}First, we show that the minimum number of clients needed for our decoding operation to be successful, i.e., the recovery threshold of {\acr}, is equal to $(2r+1)(K+T-1)+1$. 
To do so, we demonstrate in the following that the decoding process will be successful as long as $N\geq (2r+1)(K+T-1) +1$.  
As described in Section~\ref{app:multiclient_framework}, given the polynomial approximation of the sigmoid function  in~\eqref{eq:poly_approximation}, the degree of $h(z)$ in~\eqref{eq:beta} is at most $(2r+1)(K+T-1)$. 
The decoding process uses the computations from the clients as evaluation points $h(\alpha_i)$ to interpolate the polynomial $h(z)$. 
If at least $deg(h(z))+1$ evaluation results of $h(\alpha_i)$ are available, then, all of the coefficients of $h(z)$ can be evaluated. 
After $h(z)$ is recovered, the sub-gradient $\bX_i^\top \hat{g}(\bX_i \times \bw^{(t)})$ can be decoded by computing $h(\beta_i)$ for $i\in[K]$, from which the gradient $\bX^\top \hat{g}(\bX \times \bw^{(t)})$ from \eqref{eq:agg} can be computed. 
Hence, the recovery threshold of {\acr} is  $(2r+1)(K+T-1) +1$, as long as $N\geq (2r+1)(K+T-1) +1$, the protocol can correctly decode the gradient using the local evaluations of the clients, and the decoding process will be successful. 
Since the decoding operations are performed using a secure MPC protocol, throughout the decoding process, the clients only learn a secret share of the gradient and not its actual value. 
%
%
Next, we consider the update equation in \eqref{eq:apprx_grad1multi} and prove its convergence to $\mathbf{w}^*$.
As described in Section~\ref{app:multiclient_framework}, after decoding the gradient, the clients carry out a secure truncation protocol to multiply $\bX^\top (\hat{g}(\bX \times \bw^{(t)})-\mathbf{y})$ with parameter $\frac{\eta}{m}$ to update the model as in~\eqref{eq:apprx_grad1multi}.
The update equation from \eqref{eq:apprx_grad1multi} can then be represented by
\begin{align}
    \bw^{(t+1)}  &=   \bw^{(t)}\!-\eta \big(\frac{1}{m}{\mathbf{X}}^{\top}(\hat{g}({\mathbf{X}} \times {\mathbf{w}}^{(t)}) -\mathbf{y})+\mathbf{n}^{(t)} \big). \label{eq:updateEQN_secret}\\
    &=\bw^{(t)}\!-\eta \mathbf{p}^{(t)} \label{eq:define_p}
\end{align}
where $\mathbf{n}^{(t)}$ represents the quantization noise introduced by the secure multi-party truncation protocol ~\cite{catrina2010secure}, and $\mathbf{p}^{(t)} \triangleq \frac{1}{m}{\mathbf{X}}^{\top}(\hat{g}({\mathbf{X}} \times {\mathbf{w}}^{(t)}) -\mathbf{y})+\mathbf{n}^{(t)}$. 
From~\cite{catrina2010secure}, $\mathbf{n}^{(t)}$ has zero mean and bounded variance, i.e., $\mathbb{E}_{\bnt}[\mathbf{n}^{(t)}]=0$ and $\mathbb{E}_{\bnt}\big[\| \mathbf{n}^{(t)}  \|^{2}_{2} \big] \leq \frac{d2^{2(k_1-1)}}{m^2} \triangleq \sigma^2$ where $\| \cdot \|_{2}$ is the $l_2$ norm and $k_1$ is the truncation parameter described in Section~\ref{app:multiclient_framework}.

Next, we show that $\mathbf{p}^{(t)}$ is an unbiased estimator of the true gradient, $\nabla C(\bw^{(t)})=\frac{1}{m} {\mathbf{X}}^{\top} (g({\mathbf{X}} \times {\mathbf{w}}^{(t)}) -\mathbf{y})$, and its variance is bounded by $\sigma^2$ with sufficiently large $r$.
From $\mathbb{E}_{\bnt}[\mathbf{n}^{(t)}]=0$, we obtain
\begin{equation}\label{eq:delta}
    \mathbb{E}_{\bnt}[\mathbf{p}^{(t)}] - \nabla C(\bw^{(t)}) =\frac{1}{m}\bX^{\top} \big( \hat{g}({\bX}\times {\bw}^{(t)}) - g({\bX}\times {\bw}^{(t)}) \big).  
\end{equation}
From the Weierstrass approximation theorem~\cite{brinkhuis2011optimization}, for any $\epsilon>0$, there exists a polynomial that approximates the sigmoid arbitrarily well, i.e., $|\hat{g}(x)-g(x)|\leq\epsilon$ for all $x$ in the constrained interval.  
Hence, as there exists a polynomial making the norm of~\eqref{eq:delta} arbitrarily small, $\mathbb{E}_{\bnt}[\mathbf{p}^{(t)}]=\nabla C(\bw^{(t)})$ and $\mathbb{E}_{\bnt}\big[\| \mathbf{p}^{(t)} - \mathbb{E}_{\bnt}[\mathbf{p}^{(t)}] \|^{2}_{2} \big] = \mathbb{E}_{\bnt}\big[\| \mathbf{n}^{(t)}  \|^{2}_{2} \big] \leq \sigma^2$. 


Next, we consider the update equation in~\eqref{eq:define_p} and prove its convergence to $\bw^{*}$. From the $L$-Lipschitz continuity of $\nabla C({\bw})$ 
(Theorem 2.1.5 of~\cite{book_Nesterov}), we have
\begin{align}
    C(\bw^{(t+1)}) 
    &\!\leq \!C(\bw^{(t)}) \!+\! \langle \nabla C(\bw^{(t)}), \bw^{(t+1)}\!-\!\bw^{(t)}\rangle 
    \!+\! \frac{L}{2}{\| \bw^{(t+1)}\!-\!\bw^{(t)}\! \|}^2  \notag\\
    &\!\leq\! C( \bw^{(t)} )\!-\!\eta \langle \gC ( \bw^{(t)} ), \mathbf{p}^{(t)} \rangle + \frac{L\eta^2}{2}{\| \mathbf{p}^{(t)} \|}^2, \label{eq:Lipschitz_ineq}
\end{align} 
where $\langle ,\cdot, \rangle$ is the inner product. For a cross entropy loss $C(\bw)$, the Lipschitz
constant $L$ is equal to the largest eigenvalue of the Hessian $\nabla^2C(\bw)$ for all $\bw$, and is given by $L=\frac{1}{4}\| {\bX} \|^{2}_{2}$.
By taking the expectation with respect to the quantization noise $\bnt$ on both sides in~\eqref{eq:Lipschitz_ineq}, we have
\begin{align}
    \mathbb{E}_{\bnt}\big[ C(\bw^{(t+1)}) \big] 
    &\leq C( \bw^{(t)} ) - \eta \| \gC ( \bw^{(t)} )\|^2 +\frac{L\eta^2}{2} \big( \| \gC ( \bw^{(t)} )\|^2 + \sigma^2 \big) \label{eq:fisrt_ineq} \\
    &\leq C( \bw^{(t)} ) - \eta \big(1-\frac{L\eta}{2}\big)\| \gC ( \bw^{(t)} )\|^2+\frac{L\eta^2\sigma^2}{2} \notag \\
    &\leq C( \bw^{(t)} ) - \frac{\eta}{2}\| \gC ( \bw^{(t)} )\|^2\!+\!\frac{\eta\sigma^2}{2} \label{eqn1} \\
    &\leq C(\bw^{*})\!+\!\langle \gC ( \bw^{(t)} ), \bw^{(t)} \!-\! \bw^{*}\rangle 
    - \frac{\eta}{2} \| \gC ( \bw^{(t)} )\|^2\!+\!\frac{\eta\sigma^2}{2} \label{eqn2}\\
    &\leq C(\bw^{*})+ \langle\mathbb{E}_{\bnt}[\mathbf{p}^{(t)}],\bw^{(t)} - \bw^{*}\rangle - \frac{\eta}{2} \mathbb{E}_{\bnt}\| \mathbf{p}^{(t)} )\|^2+\eta\sigma^2 \label{eqn3}\\
    &= C(\bw^{*}) +\eta\sigma^2 +\mathbb{E}_{\bnt}\Big[ \langle \mathbf{p}^{(t)},\bw^{(t)}-\bw^{*} \rangle-\frac{\eta}{2}\| \mathbf{p}^{(t)} )\|^2\Big] \notag \\
    &=\! C(\bw^{*}) +\eta\sigma^2 + \frac{1}{2\eta}\big( \| \bw^{(t)} - \bw^{*} \|^2 -\mathbb{E}_{\bnt}\| \bw^{(t+1)} - \bw^{*} )\|^2\big) \label{eqn4}
\end{align}
where~\eqref{eq:fisrt_ineq} and~\eqref{eqn3} hold since  $\mathbb{E}_{\bnt}[\mathbf{p}^{(t)}]=\nabla C(\bw^{(t)})$ and $\mathbb{E}_{\bnt}\big[\| \mathbf{p}^{(t)}  - \nabla C(\bw^{(t)}) \|^{2}_{2} \big] \leq \sigma^2$,~\eqref{eqn1} follows from $L\eta\leq1$,~\eqref{eqn2} follows from the convexity of $C$, and \eqref{eqn4} follows from $\mathbf{p}^{(t)}=-\frac{1}{\eta}(\bw^{(t+1)}-\bw^{(t)})$.

By taking the expectation on both sides in \eqref{eqn4} with respect to the joint distribution of all random variables $\mathbf{n}^{(0)},\ldots,\mathbf{n}^{(J-1)}$ where $J$ denotes the total number of iterations, we have
\begin{equation} \label{eqn5}
    \mathbb{E}\big[ C(\bw^{(t+1)}) \big] - C(\bw^{*})
    \leq \frac{1}{2\eta}\big( \mathbb{E}\| \bw^{(t)} - \bw^{*} \|^2 -\mathbb{E}\| \bw^{(t+1)} - \bw^{*} )\|^2\big) +\eta\sigma^2.
\end{equation}

Summing both sides of the inequality in~\eqref{eqn5} for $t=0,\ldots,J-1$, we find that,
\begin{align*}
    \sum_{t=0}^{J-1}\Big(\mathbb{E}\big[ &C(\bw^{(t+1)})\big]\!-\!C(\bw^{*})\Big) 
    \leq \frac{\| \bw^{(0)} - \bw^{*} \|^2}{2\eta}\!+\!J\eta\sigma^2.
\end{align*}
Finally, since $C$ is convex, we observe that, 
\begin{align*}
    \mathbb{E}\Big[ C\big( \frac{1}{J}\sum_{t=0}^{J}\bw^{(t)}\big)\Big]\!-\!C(\bw^{*})
    &\leq \frac{1}{J} \sum_{t=0}^{J-1}\Big(\mathbb{E}\big[ C(\bw^{(t+1)})\big]\!-\!C(\bw^{*})\Big) \notag \\
    &\leq \frac{\| \bw^{(0)} - \bw^{*} \|^2}{2\eta J} + \eta\sigma^2
\end{align*}
which completes the proof of convergence.

\subsection{Details of the Multi-Party Computation (MPC) Implementation}~\label{MPC}
\hspace{-0.1cm}We consider two well-known MPC protocols, the notable BGW protocol from \cite{ben1988completeness}, and the more recent, efficient MPC protocol from \cite{beerliova2008perfectly,damgaard2007scalable}. 
Both protocols allow the computation of any polynomial function in a privacy-preserving manner by untrusted parties. Computations are carried out over the secret shares, and at the end, parties only learn a secret share of the actual result. Any collusions between up to $T = \lfloor \frac{N-1}{2}\rfloor$ out of $N$ parties do not reveal information (in an information-theoretic sense) about the input variables. 
The latter protocol is more efficient in terms of the communication cost between the parties, which scales linearly with respect to the number of parties, whereas for the former protocol this cost is quadratic. As a trade-off, it requires a considerable amount of offline computations and higher storage cost for creating and secret sharing the random variables used in the protocol.

For creating secret shares, we utilize Shamir's $T$-out-of-$N$ secret sharing  \cite{shamir1979share}. This scheme embeds a secret $a$ in a degree $T$ polynomial $h(\xi) = a + \xi v_1, \ldots, \xi^T v_T $ where $v_i$, $i\in [T]$ are uniformly random variables. 
Client $i\in [N]$ then  receives a secret share of $a$, denoted by $h(i) = [a]_i$. 
This keeps $a$ private against any collusions between up to any $T$ parties.  
The specific computations are then carried out as follows.

\vspace{0.1cm}
\noindent{\bf Addition.}  
In order to perform a secure addition $a+b$, clients locally add their secret shares $[a]_i+[b]_i$. The resulting value is a secret share of the original summation $a+b$. 
This step requires no communication. 

\vspace{0.1cm}
{\color{black}
\noindent{\bf Multiplication-by-a-constant.}
For performing a secure multiplication $ac$ where $c$ is a publicly-known constant, clients locally multiply their secret share $[a]_i$ with $c$. The resulting value is a secret share of the desired multiplication $ac$. 
This step requires no communication.
}

\vspace{0.1cm}
\noindent{\bf Multiplication.} 
For performing a secure multiplication $ab$, the two protocols differ in their execution. In the BGW protocol, each client initially multiplies its secret shares $[a]_i$, $[b]_i$ locally to obtain $[a]_i[b]_i$. The clients will then be holding a secret share of $ab$, however, the corresponding polynomial now has degree $2T$. This may in turn cause the degree of the polynomial to increase excessively as more multiplication operations are evaluated. To alleviate this problem, in the next phase, clients carry out a degree reduction step to create new shares corresponding to a polynomial of degree $T$.  
The communication overhead of this protocol is $O(N^2)$. 

The protocol from \cite{beerliova2008perfectly}, on the other hand, leverages offline computations to speed up the communication phase. In particular, a random variable $\rho$ is created offline and secret shared with the clients twice using two random polynomials with degrees $T$ and $2T$, respectively. The secret shares corresponding to the degree $T$ polynomial are denoted by $[\rho]_{T, i}$, whereas the secret shares for the degree $2T$ polynomial are denoted by  $[\rho]_{2T, i}$ for clients $i\in[N]$. 
In the online phase, client $i\in[N]$ locally computes the multiplication $[a]_i [b]_i$, after which each client will be holding a secret share of the multiplication $ab$. The resulting polynomial has degree $2T$. Then, each client locally computes $ [a]_i [b]_i - [\rho]_{2T, i}$, which corresponds to a secret share of $ab-\rho$ embedded in a degree $2T$ polynomial. Clients then broadcast their individual computations to others, after which each client computes $ab-\rho$. Note that the privacy of the computation $ab$ is still protected since clients do not know the actual value of $ab$, but instead its masked version $ab-\rho$. Then, each client locally computes $ab-\rho+ [\rho]_{T, i}$. As a result, variable $\rho$ cancels out, and clients obtain a secret share of the multiplication $ab$ embedded in a degree $T$ polynomial. 
This protocol requires only $O(N)$ broadcasts and therefore is more efficient than the previous algorithm. On the other hand, it requires an offline computation phase and higher storage overhead. 
For the details, we refer to \cite{beerliova2008perfectly, beerliova2008efficient}.  

\begin{remark}\normalfont
The secure MPC computations during the encoding, decoding, and model update phases of {\acr} only use addition and  multiplication-by-a-constant operations, instead of the expensive multiplication operation, as  $\{\alpha_i\}_{i\in[N]}$ and $\{\beta_k\}_{k\in[K+T]}$ are publicly known constants for all clients.
\end{remark}

\subsection{Details of the Optimized Baseline Protocols}~\label{app:opt_baseline}
\hspace{-0.12cm}In a naive implementation of our multi-client problem setting, 
both baseline protocols would utilize Shamir's secret sharing scheme where the quantized dataset $\mathbf{X} =[\mathbf{X}_1^\top, \ldots, \mathbf{X}_N^\top]^\top$ is secret shared with $N$ clients.
To do so, both baselines would follow the same secret sharing process as in {\acr}, where client $j\in[N]$ creates a degree $T$ random polynomial $h_j(z) = \bX_j + z \mathbf{R}_{j1}+ \ldots + z^T \mathbf{R}_{jT}$ where $\mathbf{R}_{ji}$ for $i\in [T]$ are i.i.d. uniformly distributed random matrices while selecting $T=\lfloor \frac{N-1}{2} \rfloor$.
By selecting $N$ distinct evaluation points $\lambda_1, \ldots, \lambda_N$ from $\mathbb{F}_p$, client $j$ would generate and send $[\bX_j]_i=h_j(\lambda_i)$ to client $i\in[N]$.
%
%
As a result, client $i\in[N]$ would be assigned a secret share of the entire dataset $\bX$, i.e, $[\bX]_i=\big[ [\bX_1]^\top_i,\ldots,[\bX_N]^\top_i\big]^\top$. 
Client $i$ would also obtain a  secret share of the labels, $[\mathbf{y}]_i$, and a secret share of the initial model, $[\bw^{(0)}]_i$, where $\mathbf{y}=[\mathbf{y}_1^\top,\ldots,\mathbf{y}_N^\top]^\top$ and $\bw^{(0)}$ is a randomly initialized model.
Then, the clients would compute the gradient and update the model from~\eqref{eq:local} within a secure MPC protocol.
This guarantees privacy against $\lfloor \frac{N-1}{2}\rfloor$ colluding workers,
but requires a computation load at each worker that is as large as processing the whole dataset 
at a single worker, leading to slow training. 

Hence, in order to provide a fair comparison with {\acr}, we optimize (speed up) the baseline protocols by partitioning the clients into subgroups of size $2T + 1$.
Clients communicate a secret share of their own datasets with the other clients in the same subgroup, instead of secret sharing it with the entire set of clients.  
Each client in subgroup $i$ receives a secret share of a partitioned dataset $\mathbf{X}_i\in\mathbb{F}^{\frac{m}{G}\times d}_p$ where $\mathbf{X} = [\mathbf{X}_1^\top \cdots \mathbf{X}_G^\top]^\top$ and $G$ is the number of subgroups. 
In other words, client $j$ in subgroup $i$ obtains a secret share $[\bX_i]_j$. 
Then, subgroup $i\in[G]$ computes the sub-gradient over the partitioned dataset, $\bX_i$, within a secure MPC protocol. 
To provide the same privacy threshold $T=\lfloor \frac{N-3}{6} \rfloor$ as Case 2 of {\acr} in Section~\ref{sec:experiments}, we set $G=3$.
This significantly reduces the total training time of the two baseline protocols (compared to the naive MPC implementation where the computation load at each client would be as high as training centrally), as the total amount of data processed at each client is equal to one third of the size of the entire dataset $\bX$.

\subsection{Algorithms}~\label{app:algorithms}
\hspace{-0.09cm}The overall procedure of {\acr} protocol is given in Algorithm~\ref{algorithm:masterless}.

\begin{algorithm*}[ht]
\footnotesize
  \caption{{\acr}}\label{algorithm:masterless} 
  \begin{algorithmic}[1]
    \INPUT{Dataset $(\bX,\mathbf{y})=((\bX_1,    \mathbf{y}_1), \ldots, (\bX_N, \mathbf{y}_N))$ distributed over $N$ clients.} \
    \OUTPUT{Model parameters $\bw^{(J)}$.} \ 
    \vspace{0.1cm}
    \FOR{client $j=1,\ldots,N$}
    
    \STATE  Secret share the individual dataset $(\bX_j, \mathbf{y}_j)$ with clients $i\in[N]$.     
    \ENDFOR
    
    \STATE  Within a secure MPC protocol, initialize the model $\bw^{(0)}$ randomly and secret share with clients $i\in[N]$.     \NEWCOMMENT{Client $i$ receives a secret share $[\bw^{(0)}]_i$ of $\bw^{(0)}$.}
    \vspace{0.1cm}
    \STATE Encode the dataset within a secure MPC  protocol, using the secret shares $[\bX_j]_i$ for $j\in [N]$, $i\in [N]$.     \NEWCOMMENT{After this step, client $i$ holds a secret share $[\widetilde{\bX}_j]_i$ of each encoded dataset $\widetilde{\bX}_j$ for $j\in [N]$.} 
    \FOR{client $i=1,\ldots,N$}
    \STATE  Gather the secret shares $[\widetilde{\bX}_i]_j$ from  clients $j\in [N]$.
    \STATE Recover the encoded dataset  $\widetilde{\bX}_i$ from the secret shares $\{[\widetilde{\bX}_i]_j\}_{j\in[N]}$. 
    \NEWCOMMENT{At the end of this step, client $i$ obtains the encoded dataset $\widetilde{\bX}_i$.} 
    \ENDFOR
    \STATE  Compute $\bX^T \mathbf{y}$ within a secure MPC  protocol using the secret shares $[\bX_j]_i$ and $[\mathbf{y}_j]_i$ for $j\in [N]$, $i\in [N]$. \NEWCOMMENT{At the end of this step, client $i$ holds a secret share $[\bX^T \mathbf{y}]_i$ of $\bX^T \mathbf{y}$.} 
    
    \vspace{0.1cm}
    \FOR{iteration $t=0,\ldots,J-1$}
    
    \STATE  Encode the model $\bw^{(t)}$ in a secure MPC  protocol using the secret shares $[\bw^{(t)}]_i$. \NEWCOMMENT{After this step, client $i$ holds  a secret share $[\widetilde{\bw}^{(t)}_j]_i$ of the encoded model $\widetilde{\bw}^{(t)}_j$ for $j\in [N]$.} 
    
    \FOR{client $i=1,\ldots,N$}
    \STATE  Gather the secret shares $[\widetilde{\bw}^{(t)}_i]_j$ from  clients $j\in [N]$.
    \STATE Recover the encoded model  $\widetilde{\bw}^{(t)}_i$ from the secret shares $\{[\widetilde{\bw}^{(t)}_i]_j\}_{j\in[N]}$. 
    \NEWCOMMENT{At the end of this step, client $i$ obtains the encoded model  $\widetilde{\bw}^{(t)}_i$.}
    \STATE  Locally compute  $f(\widetilde{\bX}_i,\widetilde{\bw}_i^{(t)})$ from    \eqref{eq:local} and secret share the result with clients $j\in [N]$. \NEWCOMMENT{Client $i$ sends a secret share $[f(\widetilde{\bX}_i,\widetilde{\bw}_i^{(t)})]_j$ of $f(\widetilde{\bX}_i,\widetilde{\bw}_i^{(t)})$  to client $j$.} 
    \ENDFOR
    
    \FOR{client $i=1,\ldots,N$}
    \STATE Locally computes $[f(\bX_k, \bw^{(t)})]_i$ for $k\in [K]$ from~\eqref{eq:local_dec}.  
    \NEWCOMMENT{After this step, client $i$ knows a secret share $[f(\bX_k, \bw^{(t)})]_i$ of $f(\bX_k, \bw^{(t)})$ for $k\in [K]$.}
    \STATE  Locally aggregate the secret shares $\{[f(\bX_k, \bw^{(t)})]_i\}_{k\in \mathcal{K}}$ to compute $[\bX^T\hat{g}(\bX\times \bw^{(t)})]_i \triangleq \sum_{k\in [K]}[f(\bX_k, \bw^{(t)})]_i$. \NEWCOMMENT{At the end of this step, client $i$ now has a secret share $[\bX^T\hat{g}(\bX\times \bw^{(t)})]_i$ of $\bX^T\hat{g}(\bX\times \bw^{(t)}) = \sum_{k\in [K]}f(\bX_k, \bw^{(t)})$.} 
    
    \STATE 
    Locally compute  $[\mathbf{X}^{\top}(\hat{g}(\mathbf{X} \times \mathbf{w}^{(t)}) -\mathbf{y})]_i \triangleq [\bX^T\hat{g}(\bX\times \bw^{(t)})]_i - [\bX^T \mathbf{y}]_i$. 
     \NEWCOMMENT{Each client now has a secret share  $[\mathbf{X}^{\top}(\hat{g}(\mathbf{X} \times \mathbf{w}^{(t)}) -\mathbf{y})]_i$ of  $\mathbf{X}^{\top}(\hat{g}(\mathbf{X} \times \mathbf{w}^{(t)}) -\mathbf{y})$.}  
    \ENDFOR
     \STATE 
    Update the model according to \eqref{eq:apprx_grad1multi} within a secure MPC  protocol using the secret shares $[\mathbf{X}^{\top}(\hat{g}(\mathbf{X} \times \mathbf{w}^{(t)}) -\mathbf{y})]_i$ and $[\bw^{(t)}]_i$ for $i\in [N]$, and by carrying out the secure truncation operation.

    \NEWCOMMENT{At the end of this step, client $i$ holds a secret share of the updated model  $[\bw^{(t+1)}]_i$.}  
    \NEWCOMMENT{Secure truncation is carried out jointly as it requires  communication between the clients.}    
    
    \ENDFOR
    
    \vspace{0.1cm}
    
    \FOR{client $j=1,\ldots,N$}
 \STATE Collect the secret shares $[\bw^{(J)}]_i$ from clients $i\in [N]$ and recover the final model $\bw^{(J)}$. 
    \ENDFOR
  \end{algorithmic}
\end{algorithm*}

\end{document}